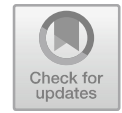

# Estimating the Optimal Number of Clusters in Categorical Data Clustering by Silhouette Coefficient

Duy-Tai Dinh[(✉)] , Tsutomu Fujinami, and Van-Nam Huynh

School of Knowledge Science, Japan Advanced Institute of Science and Technology,
1-1 Asahidai, Nomi, Ishikawa 923-1292, Japan
{taidinh,fuji,huynh}@jaist.ac.jp

**Abstract.** The problem of estimating the number of clusters (say $k$) is one of the major challenges for the partitional clustering. This paper proposes an algorithm named $k$-SCC to estimate the optimal $k$ in categorical data clustering. For the clustering step, the algorithm uses the kernel density estimation approach to define cluster centers. In addition, it uses an information-theoretic based dissimilarity to measure the distance between centers and objects in each cluster. The silhouette analysis based approach is then used to evaluate the quality of different clusterings obtained in the former step to choose the best $k$. Comparative experiments were conducted on both synthetic and real datasets to compare the performance of $k$-SCC with three other algorithms. Experimental results show that $k$-SCC outperforms the compared algorithms in determining the number of clusters for each dataset.

## 1 Introduction

Clustering is one of the important data mining techniques for discovering knowledge in unstructured multivariate and multidimensional data. The goal of clustering is to identify pattern or groups of similar objects in a given dataset. Each group, or clusters, consists of objects that are similar to one another and dissimilar to objects in other groups. Clustering has been developed in many fields with very diverse applications such as scientific data exploration, information retrieval and text mining, web analysis, marketing, medical diagnostics, computational biology and many others [2]. Clustering algorithms can be classified by type of clustering (hard and soft clustering, flat and hierarchical clustering), type of data (nominal, ordinal, interval scaled and mixed data), by clustering criterion ((probabilistic) model-based and cost-based clustering), or by regime (parametric and non-parametric clustering) [7]. Among these methods, partitional clustering, which is a kind of flat clustering, aims to discover the groupings present in the data by optimizing a specific objective function and iteratively improving the quality of the partitions [16].

$K$-means clustering [12] is the most typical representative for the group of partitional clustering algorithms. It is based on the idea of using the cluster centers (means) as representatives of each cluster. $K$-means has been widely used in many real-life applications due to their simplicity and competitive computational complexity. However, one of the limitations of $k$-means is that it can not be applied directly to categorical data, which is common in real datasets. To tackle this problem, a data transformation based method can be used to first transform categorical data into a new feature space, and then apply $k$-means to the newly transformed space to obtain the final results. However, this method has proven to be very ineffective and does not produce good clusters [16]. During the last two decades or so, several attempts have been made to remove the numeric-only limitation of the $k$-means algorithm and make it applicable to clustering for categorical data [4–6, 8, 9, 13–15, 18]. These algorithms use a similar clustering procedure to $k$-means, but differ from each other with regards to defining cluster centers and dissimilarity measures for categorical data.

In partitional clustering, the major factors that can impact the performance of this type of algorithms are choosing the initial centroids and estimating the number of clusters [16]. Particularly, in the former case, the random initialization method has been widely used in $k$-means and $k$-means-like algorithms for its simplicity. However, this method may yield different clustering results on different runs of the algorithms, and very poor clustering results may occur in some cases. In the latter case, most algorithms assume the number of clusters in advance. However, a fixed number of clusters may lead to difficulties in predicting the actual number of clusters and thus influence the interpretation of the results. Moreover, the over-estimation or under-estimation of the number of clusters will considerably affect the quality of clustering results [10]. Thus, identifying the number of clusters in a dataset is a vital task in clustering analysis. This paper focuses on solving the problem of estimating the number of clusters in categorical data clustering. More specifically, the key contributions of this paper are as follows:

- We propose a $\underline{k}$-means-like clustering algorithm based on the $\underline{s}$ilhouette analysis approach to estimate the optimal number of clusters in $\underline{c}$ategorical data $\underline{c}$lustering, namely $k$-SCC.
- We perform an extensive experiment on both synthetic and real datasets from the UCI Machine Learning Repository to evaluate the performance of the proposed $k$-SCC algorithm in terms of clustering quality based on the average silhouette values. The output of the proposed algorithm suggests the optimal number of clusters for each dataset.
- The proposed method is applied to find the optimal number of clusters for a real Sake wine dataset as a case study.

The rest of this paper is organized as follows. Section 2 gives a brief overview of the related work. Section 3 introduces preliminaries and problem statement. Section 4 shows the proposed algorithm for clustering categorical data. Section 5 describes experimental results. Finally, Sect. 6 draws a summary and outlines directions for future work.

## 2 Related Work

$K$-means clustering [12] is a well-known algorithm in partitional clustering. It starts by choosing $k$ samples (objects) as the initial centroids. Each sample in the dataset is then assigned into the nearest centroids based on a particular proximity measure, of which the most frequently used are Manhattan, Euclidean and Cosine distances. Once the clusters are formed, the centroids are updated. The algorithm iteratively performs the assignment and update steps until a convergence criterion is met. Given a dataset $D = \{x_1, \ldots, x_n\}$ with $n$ samples and $m$ features, $k$-means aims to minimize the following objective function:

$$F(U, Z) = \sum_{l=1}^{k} \sum_{i=1}^{n} \sum_{j=1}^{m} u_{il} \times d(x_{ij}, z_{lj}) \quad (1)$$

where $k$ be the number of clusters; $U = [u_{il}]$ is an $n \times k$ partition matrix that satisfies $u_{il} \in \{0,1\}$ and $\sum_{l=1}^{k} u_{il} = 1$ $(1 \leq i \leq n; 1 \leq l \leq k)$; $Z = \{Z_l, l = 1, \ldots, k\}$ is a set of cluster centers in which $Z_l$ consists of $m$ values $(z_1^l, z_2^l, \ldots, z_m^l)$, each being the mean of an attribute $j$ in cluster $Z_l$ and defined as $z_j^l = \frac{\sum_{x_i \in Z_l} x_{ij}}{|Z_l|}$; $d(\cdot, \cdot)$ is the squared Euclidean between two attribute values. $K$-means has many advantages such as easy implementation, easy interpretation and efficient computation. However, working only on numerical data prohibits some applications of $k$-means. Particularly, it can not be directly applied to categorical data which is fairly common in real datasets.

To tackle this problem, several $k$-means-like algorithms have been proposed for categorical data clustering. $K$-modes [8,9] extends $k$-means by using *modes* to define cluster centers and the simple matching dissimilarity measure to measure the distance of categorical objects. Let $x_a$ and $x_b$ be two categorical values. The simple matching distance between $x_a$ and $x_b$ is given by:

$$\delta(x_a, x_b) = \begin{cases} 0 & \text{if } x_a = x_b \\ 1 & \text{if } x_a \neq x_b \end{cases} \quad (2)$$

As $k$-means, $k$-modes is also an optimization problem and has similar runtime, benefits, and disadvantages [9].

$K$-representatives [18] applies the other approach to define cluster representative. It uses the distribution of categorical values appearing in each attribute to define the representative at that attribute. The dissimilarity between each object and representative is measured based on the multiplication of relative frequencies of categorical values within the cluster and the simple matching measure between categorical values. $K$-centers [4] uses the kernel density estimation method to define the center of a categorical cluster, called the probabilistic center. It incorporates a built-in feature weighting in which each attribute is automatically assigned with a weight to measure its contribution for the clusters. More recently, several works have been proposed to improve clustering results or deal with the problem of missing values in categorical data [5,6,13,15]. In [15], three modifications of $k$-representatives were introduced. The first version (Modified-1) replaces the simple matching measure with an information-theoretic based dissimilarity measure. The second version (Modified-2) uses the new dissimilarity measure with the concept of cluster centers proposed in $k$-centers [4] to form clusters. The third version (Modified-3) uses the dissimilarity as in Modified-2 and incorporates with a modified representative of cluster centers using the kernel density estimation method. Modified-3 was shown to be more efficient than the former two versions in terms of clustering quality.

The silhouette coefficient [17] is an internal measure for cluster validation. It considers both the intra-cluster and inter-cluster distances. For each data sample $x_i$, the average of the distances to all samples in the same cluster is first calculated and set to $a(x_i)$. For each cluster that does not contain this sample, the average distance of $x_i$ to all samples in each cluster is calculated. Then the smallest of these distances is taken and set to $b(x_i)$. The two values $a(x_i)$ and $b(x_i)$ are used to estimate the silhouette coefficient of $x_i$. The average of all the silhouettes in the dataset is called the average silhouette for all samples in the dataset. The average silhouette can be used to evaluate the quality of a clustering. In addition, the optimal cluster number can be determined by maximizing the value of this index. Several works based on silhouette coefficient have been proposed to estimate the number of clusters in numerical data clustering [1,20]. To the best of our knowledge, no prior work has been conducted on estimating the number of clusters in categorical data clustering using the silhouette coefficient. The next section shows preliminary definitions of the proposed framework.

## 3   Preliminaries

Let $n$ and $m$ be the number of objects and attributes in a categorical dataset, respectively. A categorical dataset $D$ is an $n \times m$ matrix ($n \gg m$) in which each element at position $(i, j)$ ($1 \leq i \leq n$, $1 \leq j \leq m$) stores the value of an object $x_i$ at the $j^{th}$ attribute. A categorical object $x_i \in D$ ($1 \leq i \leq n$) is a tuple of $m$ values $x_i = (x_{i1}, x_{i2}, \ldots, x_{im}) \in A_1 \times A_2 \times \cdots \times A_m$, where $A_j$ is a categorical attribute which is characterized by a finite domain $\mathcal{O}_j$ such that $\text{DOM}(A_j) = |\mathcal{O}_j|$ ($> 1$) discrete values. In addition, a category in $\mathcal{O}_j$ is denoted by $o_{ij}$ ($1 \leq i \leq |\mathcal{O}_j|$). For example, Table 1 shows a categorical dataset that contains ten objects with six categorical attributes.

**Table 1.** A categorical dataset

| Obj | Attr | | | | | |
|---|---|---|---|---|---|---|
| | $A_1$ | $A_2$ | $A_3$ | $A_4$ | $A_5$ | $A_6$ |
| $x_1$ | $a_1$ | $d_2$ | $b_3$ | $e_4$ | $a_5$ | $c_6$ |
| $x_2$ | $d_1$ | $a_2$ | $a_3$ | $b_4$ | $c_5$ | $a_6$ |
| $x_3$ | $d_1$ | $d_2$ | $d_3$ | $c_4$ | $c_5$ | $a_6$ |
| $x_4$ | $b_1$ | $e_2$ | $c_3$ | $e_4$ | $a_5$ | $c_6$ |
| $x_5$ | $a_1$ | $d_2$ | $a_3$ | $a_4$ | $a_5$ | $e_6$ |
| $x_6$ | $a_1$ | $a_2$ | $c_3$ | $e_4$ | $a_5$ | $c_6$ |
| $x_7$ | $b_1$ | $e_2$ | $e_3$ | $e_4$ | $a_5$ | $c_6$ |
| $x_8$ | $d_1$ | $c_2$ | $d_3$ | $e_4$ | $a_5$ | $c_6$ |
| $x_9$ | $d_1$ | $c_2$ | $d_3$ | $e_4$ | $a_5$ | $c_6$ |
| $x_{10}$ | $d_1$ | $d_2$ | $b_3$ | $b_4$ | $c_5$ | $a_6$ |

**Definition 1 (Clusters).** Given a categorical dataset $D = \{x_1, \ldots, x_n\}$, let $C = \{C_1, C_2, \ldots, C_k\}$ be a set of $k$ disjoint subsets that contain the indices of objects in $D$. These subsets are called clusters if they satisfy the two conditions: $C_l \cap C_{l'} = \emptyset$ for all $l \neq l'$ and $\bigcup_{l=1}^{k} C_l = D$. The number of objects in cluster $C_l$ is denoted by $n_l$.

**Definition 2 (Relative frequency).** Let there be a cluster $C_l$, the relative frequency of a category $o_{ij}^l$ ($1 \leq i \leq n_l, 1 \leq j \leq m$) occurring in $C_l$ at the $j^{th}$ attribute is defined as:

$$f_l(o_{ij}^l) = \frac{\#_l(o_{ij}^l)}{n_l} \quad (3)$$

with $\#_l(o_{ij}^l)$ being the number of $o_{ij}^l$ appearing in cluster $C_l$ at the $j^{th}$ attribute.

The relative frequency of a category $o_{ij}$ occurring at the $j^{th}$ attribute of dataset $D$ is defined as:

$$f(o_{ij}) = \frac{\#(o_{ij})}{n} \quad (4)$$

and $\#(o_{ij})$ denotes the number of $o_{ij}$ occurring in dataset $D$ at the $j^{th}$ attribute.

In partitional clustering, each cluster is represented by its center. From a statistical perspective, the cluster center of a numeric cluster is the expectation of a continuous random variable associated with the data, based on the assumption that the variable follows a Gaussian distribution [4]. Following this perspective, the center of a categorical cluster can be estimated by using the kernel density estimation method, called the probabilistic center [4–6,13,15]. In particular, the center of a categorical cluster $C_l$ is denoted as $Z_l = \{z_j^l\}_{j=1}^{m}$ where the $j^{th}$ element being a vector in the probability space and $\mathcal{O}_j^l$, $P_j^l$ (in Definition 5) serves as the sample space and the probability measure defined on the Borel set of the sample space with regard to data subset $C_l$, respectively.

**Definition 3 (Kernel density estimation method).** Let there be a cluster $C_l$, let $X_j^l$ be the random variable associated with observations $x_i$ ($1 \leq i \leq n_l$) occurring in cluster $C_l$ at the $j^{th}$ attribute and $p(X_j^l)$ be its probability of density. Let $\mathcal{O}_j^l$ denote the set of categories occurring at the $j^{th}$ attribute of $C_l$ such that $\mathcal{O}_j^l = \bigcup_{i=1}^{n_l} x_{ij}$ and $\lambda_l \in [0,1]$ be the unique smoothing bandwidth for cluster $C_l$. For each value $o_{ij}^l$ in $\mathcal{O}_j^l$ ($1 \leq i \leq n_l$), the variation on Aitchison & Aitken's kernel function is denoted and defined as:

$$K(X_j^l, o_{ij}^l, \lambda_l) = \begin{cases} 1 - \frac{|\mathcal{O}_j^l|-1}{|\mathcal{O}_j^l|}\lambda_l & \text{if } X_j^l = o_{ij}^l \\ \frac{1}{|\mathcal{O}_j^l|}\lambda_l & \text{otherwise} \end{cases} \quad (5)$$

Note that the above kernel function is estimated in terms of the cardinality of the subdomain $\mathcal{O}_j^l$ of cluster $C_j$. Let $\hat{p}(X_j^l, \lambda_l, C_l)$ be the kernel estimator of $p(X_j^l)$. It is defined as:

$$\hat{p}(X_j^l, \lambda_l, C_l) = \sum_{o_{ij}^l \in \mathcal{O}_j^l} f_l(o_{ij}^l) K(X_j^l, o_{ij}^l, \lambda_l) \quad (6)$$

**Definition 4 (Smoothing bandwidth parameter).** Let there be a cluster $C_l$, a smoothing bandwidth parameter using the least square cross-validation is used to minimize the total error of the resulting estimation over data objects in this cluster. The optimal smoothing parameter for $C_l$ is denoted and defined as:

$$\lambda_l = \frac{1}{(n_l-1)} \frac{\sum_{j=1}^{m}(1 - \sum_{o_{ij}^l \in \mathcal{O}_j^l}[f_l(o_{ij}^l)]^2)}{\sum_{j=1}^{m}(\sum_{o_{ij}^l \in \mathcal{O}_j^l}[f_l(o_{ij}^l)]^2 - \frac{1}{|\mathcal{O}_j^l|})} \quad (7)$$

**Definition 5 (Probabilistic Center of a Categorical cluster).** Let there be a cluster $C_l = \{x_1, x_2, \ldots, x_{n_l}\}$ where $x_i = (x_{i1}, x_{i2}, \ldots, x_{im})$ ($1 \leq i \leq n_l$). Let $\mathcal{O}_j^l$ be a set of categories occurring at the $j^{th}$ attribute in cluster $C_l$. The center of $C_l$ is denoted and defined as:

$$Z_l = \{z_1^l, z_2^l, \ldots, z_m^l\} \quad (8)$$

where the value at $j^{th}$ element of $Z_l$ is a probability distribution on $\mathcal{O}_j^l$ estimated by a kernel density estimation method using Eq. (6) and is defined as:

$$z_j^l = [P_j^l(o_{1j}^l), P_j^l(o_{2j}^l), \ldots, P_j^l(o_{|\mathcal{O}_j^l|j}^l)] \quad (9)$$

and the value of each category $o_{ij}^l$ ($1 \leq i \leq |\mathcal{O}_j^l|$) is measured by using Eqs. (3), (5) and (6) as follows:

$$P_j^l(o_{ij}^l) = \begin{cases} \lambda_l \frac{1}{|\mathcal{O}_j^l|} + (1-\lambda_l)f_l(o_{ij}^l) & \text{if } o_{ij}^l \in \mathcal{O}_j^l \\ 0 & \text{otherwise} \end{cases} \quad (10)$$

The probabilistic estimation of a category shown in Eq. (10) can be seen as a Bayes-type probability estimator where the uniform probability ($\frac{1}{|\mathcal{O}_j^l|}$) is as a prior and the frequency estimator ($f_l(o_{ij}^l)$) is as the posterior [4].

Methods for quantifying the distance between categorical objects and cluster centers have been extensively studied in recent years [3,8,11,15,19]. In general, similarity measures which allow the comparison of values in an attribute can be classified into three types. The first type assigns possible values that are different than zero for the similarity in which a match occurs, while a value of zero is assigned for the similarity in which a value of mismatch occurs. The second type assigns a value of one for the similarity in which a value match occurs, while possible value different than one is assigned for the similarity in which a value of mismatch occurs. The last type defines different values when matching and mismatching occur [19]. The Lin similarity [11] falls into the last type. It defines similarity as the relationship between the common and different information component from the view of information theory, which uses the logarithmic function to calculate the real value in such a way that less frequent words have a higher information gain. Based on Lin similarity, an information-theoretic based dissimilarity measure has been proposed for categorical data [3,15], which is used in this paper to determine the distance between a categorical object and its cluster center.

**Definition 6 (Dissimilarity between two categories).** The similarity of two categories $o_{ij}$ and $o_{i'j}$ occurring in two objects $x_i$ and $x_{i'}$ at the $j^{th}$ attribute is defined as:

$$\text{sim}_j(o_{ij}, o_{i'j}) = \frac{2 \log f(o_{ij}, o_{i'j})}{\log f(o_{ij}) + \log f(o_{i'j})} \quad (11)$$

with $f(o_{ij}, o_{i'j}) = \frac{\#(o_{ij}, o_{i'j})}{|D|}$ be the relative frequency of two categorical objects in dataset $D$ that receive the value belonging to $\{o_{ij}, o_{i'j}\}$ at the $j^{th}$ attribute. The dissimilarity between $o_{ij}$ and $o_{i'j}$ at the $j^{th}$ attribute is measured as:

$$\text{dsim}_j(o_{ij}, o_{i'j}) = 1 - \text{sim}_j(o_{ij}, o_{i'j}) = 1 - \frac{2 \log f(o_{ij}, o_{i'j})}{\log f(o_{ij}) + \log f(o_{i'j})} \quad (12)$$

**Definition 7 (Dissimilarity between two categorical objects).** Let there be two categorical objects $x_i = (x_{i1}, x_{i2}, \ldots, x_{im})$ and $x_{i'} = (x_{i'1}, x_{i'2}, \ldots, x_{i'm})$, the dissimilarity of $x_i$ and $x_{i'}$ is defined as:

$$\text{dsim}(x_i, x_{i'}) = \sum_{j=1}^{m} \text{dsim}_j(x_{ij}, x_{i'j}) \quad (13)$$

Table 2 shows the pairwise distance matrix of all data objects shown in Table 1. This is a symmetric matrix where $\text{dsim}(x_i, x_{i'}) = \text{dsim}(x_{i'}, x_i)$ and $\text{dsim}(x_i, x_i) = 0$ ($1 \leq i, i' \leq n$). Intuitively, the dissimilarity of two objects is zero if they are identical. The upper bound dissimilarity value of them is exactly the number of attributes, where categorical values at each attribute of two objects are different.

**Table 2.** Dissimilarity matrix of categorical objects

|    | 1      | 2      | 3      | 4      | 5      | 6      | 7      | 8      | 9      | 10     |
|----|--------|--------|--------|--------|--------|--------|--------|--------|--------|--------|
| 1  | **0.0000** | 4.3518 | 3.7898 | 1.4107 | 2.308  | 1.4107 | 1.3645 | 1.7617 | 1.7617 | 3.3256 |
| 2  | 4.3518 | **0.0000** | 1.4872 | 4.2182 | 3.1476 | 3.7875 | 4.172  | 3.6046 | 3.6046 | 1.0262 |
| 3  | 3.7898 | 1.4872 | **0.0000** | 4.4165 | 2.9759 | 4.4165 | 4.3866 | 3.2191 | 3.2191 | 0.8917 |
| 4  | 1.4107 | 4.2182 | 4.4165 | **0.0000** | 2.9497 | 0.8614 | 0.3845 | 1.6281 | 1.6281 | 4.383  |
| 5  | 2.308  | 3.1476 | 2.9759 | 2.9497 | **0.0000** | 2.5191 | 2.9035 | 3.2858 | 3.2858 | 2.9827 |
| 6  | 1.4107 | 3.7875 | 4.4165 | 0.8614 | 2.5191 | **0.0000** | 1.2458 | 1.6281 | 1.6281 | 4.383  |
| 7  | 1.3645 | 4.172  | 4.3866 | 0.3845 | 2.9035 | 1.2458 | **0.0000** | 1.5983 | 1.5983 | 4.3368 |
| 8  | 1.7617 | 3.6046 | 3.2191 | 1.6281 | 3.2858 | 1.6281 | 1.5983 | **0.0000** | 0.0000 | 3.7694 |
| 9  | 1.7617 | 3.6046 | 3.2191 | 1.6281 | 3.2858 | 1.6281 | 1.5983 | 0.0000 | **0.0000** | 3.7694 |
| 10 | 3.3256 | 1.0262 | 0.8917 | 4.383  | 2.9827 | 4.383  | 4.3368 | 3.7694 | 3.7694 | **0.0000** |

**Definition 8 (Distance between an object and cluster center).** Let there be a cluster $C_l$ with its center $Z_l = \{z_1^l, z_2^l, \ldots, z_m^l\}$ and a categorical object $x_i = (x_{i1}, x_{i2}, \ldots, x_{im})$. Let $\mathcal{O}_j^l$ be a set of categories appearing at the $j^{th}$ attribute of $z_j^l$. The dissimilarity between $x_i$ and $Z_l$ at the $j^{th}$ attribute is measured by accumulating the probability distribution on $\mathcal{O}_j^l$ and the dissimilarity between $j^{th}$ component $x_{ij}$ of the object $x_i$ and the $j^{th}$ component $z_j^l$ of the center $Z_l$, which is formulated as:

$$\text{dis}_j(x_i, Z_l) = \sum_{o_{ij}^l \in \mathcal{O}_j^l} P_j^l(o_{ij}^l) \text{dsim}_j(x_{ij}, o_{ij}^l) \tag{14}$$

The distance between $x_i$ and cluster center $Z_l$ is then measured as:

$$\text{dis}(x_i, Z_l) = \sum_{j=1}^{m} \text{dis}_j(x_i, Z_l) \tag{15}$$

Based on the above distance measure, the categorical data clustering algorithm aims to minimize the following optimization function:

$$F(U, Z) = \sum_{l=1}^{k} \sum_{i=1}^{n} u_{i,l} \times dis(x_i, Z_l) \tag{16}$$

subject to

$$\begin{cases} \sum_{l=1}^{k} u_{i,l} = 1 & 1 \leq i \leq n \\ u_{i,l} \in \{0, 1\} & 1 \leq l \leq k, \ 1 \leq i \leq n \end{cases} \tag{17}$$

where $U = [u_{i,l}]_{n \times k}$ is the partition matrix in which $u_{i,l}$ takes value 1 if object $x_i$ is in cluster $C_l$ and 0 otherwise.

**Definition 9 (Silhouette value of a categorical object).** Let there be a cluster $C_l$, let $x_i$ be a categorical object in $C_l$ ($1 \leq i \leq n_l$). Let $a(x_i)$ be the average distance of the $x_i$ to all other members of the same cluster $C_l$. Let $C_{l'}$ be some cluster other than $C_l$ and let $d(x_i, C_{l'})$ be the average distance of the $x_i$ to all members of $C_{l'}$. Compute $d(x_i, C_{l'})$ for all clusters $C_{l'}$ other than $C_l$ and let $b(x_i) = \min_{C_{l'} \neq C_l} d(x_i, C_{l'})$. If cluster $C_\alpha$ ($1 \leq \alpha \leq k$) satisfies the condition that $b(x_i) = d(x_i, C_\alpha)$ then $C_\alpha$ is called the neighbor of $x_i$ and is considered as the second-best cluster for the $x_i$. The silhouette value of $x_i$ is denoted and defined as:

$$s(x_i) = \frac{b(x_i) - a(x_i)}{\max\{a(x_i), b(x_i)\}} \tag{18}$$

The Eq. (18) shows that the silhouette value is between $-1$ and 1. A large positive value of $s(x_i)$, i.e. $a(x_i)$ approximately equals to zero, indicates that the *within* dissimilarity $a(x_i)$ is much smaller than the smallest *between* dissimilarity $b(x_i)$ and thus $x_i$ is well-clustered. A large negative value of $s(x_i)$, i.e. $b(x_i)$ approximately equals to zero, indicates that $x_i$ is poor-clustered. If $s(x_i)$ is about zero, i.e. $a(x_i)$ approximately equals to $b(x_i)$, indicate that $x_i$ lies between two clusters. In general, $s(x_i)$ measures how well object $x_i$ has been classified into cluster $C_l$.

**Definition 10 (Average silhouette value).** Let there be a set of $k$ clusters $C = \{C_1, C_2, \ldots, C_k\}$, the average of all silhouette values for all categorical objects in the dataset is called the *average silhouette value* and is defined as:

$$s_k = \frac{\sum_{i=1}^n s(x_i)}{n} \tag{19}$$

In this paper, the problem of estimating the number of clusters in categorical data clustering is to find the optimal $k$ to maximize the average silhouette value shown in Eq. (19).

The next section proposes an algorithm named $k$-SCC for estimating the number of clusters in categorical data clustering.

## 4 The Proposed $k$-SCC Algorithm

The proposed $k$-SCC algorithm is based on the general framework depicted in Fig. 1. According to this model, the proposed algorithm partitions a categorical data into $k$ groups and then computes the average silhouette value for the current iteration. This process works in the same manner for every index $k$ in the range of predefined minimum and maximum number of clusters. Finally, the index that yields the largest value of the average silhouette is selected as the optimal number of clusters for that categorical data.

The pseudo code of the $k$-SCC algorithm is shown in Algorithm 1. The input of this algorithm is a categorical data $D$ and two predefined minimum and maximum numbers of clusters, denoted by $k_{\min}$ and $k_{\max}$, respectively. By default, $k_{\min}$ and $k_{\max}$ are in range $[2, n-1]$, where $n$ is the number of objects in the given dataset. A set of average silhouettes, namely *SilSet*, is used to keep the average silhouette at every iteration of the algorithm (line 1). The algorithm then computes the dissimilarity matrix that contains dissimilarity of pairs of

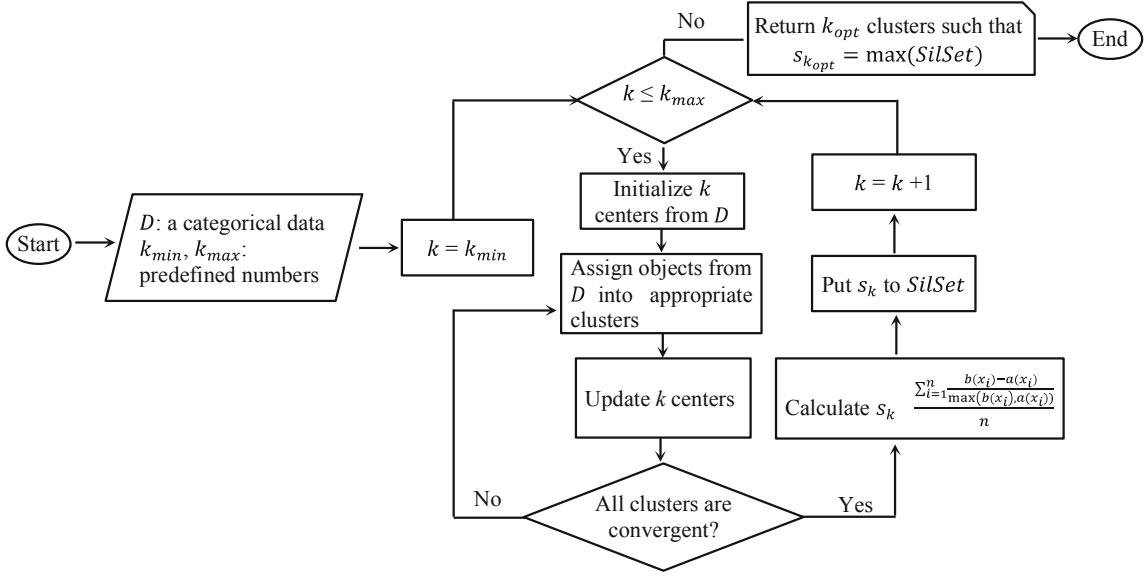

**Fig. 1.** The flowchart of $k$-SCC algorithm

---

**Algorithm 1.** THE $k$-SSC ALGORITHM

**input** : $D$: a categorical dataset, $k_{\min}$, $k_{\max}$: the minimum and maximum value of $k$.
**output**: $k$ clusters of $D$

1  SilSet $\leftarrow \emptyset$, $k = k_{\min}$
2  Compute the dissimilarity matrix of all categorical objects in $D$ using Eq. (13)
3  **while** $k \leq k_{\max}$ **do**
4  $\quad$ Initialize $k$ cluster centers $Z^{(0)} = \{Z_1^{(0)}, \ldots, Z_k^{(0)}\}$
5  $\quad$ $U \leftarrow \emptyset$, $t = 0$
6  $\quad$ **while** Partitions are not convergent **do**
7  $\quad\quad$ Keep $Z^{(t)}$ fixed, generate $U^{(t)}$ to minimize the distances between objects and cluster centers using Eq. (15)
8  $\quad\quad$ Keep $U^{(t)}$ fixed, update $Z^{(t)}$ using Eq. (8)
9  $\quad\quad$ $t = t + 1$
10 $\quad$ Compute the average silhouette value $s_k$ using Eq. (19)
11 $\quad$ SilSet $\leftarrow s_k$, $k = k + 1$
12 **return** $k_{opt}$ clusters such that $s_{k_{opt}} = \max(SilSet)$;

---

objects in $D$ based on the Eq. (13) (line 2). This matrix is used later for computing the *within* and *between* dissimilarities in average silhouette values. For every index $k$ in the range of $k_{\min}$ and $k_{\max}$, $k$-SCC performs assignment and update steps to partition categorical objects into $k$ clusters (lines 4 to 9). In the initial step, $k$-SCC initiates $k$ cluster centers by randomly selecting $k$ objects from $D$, one for each cluster (line 4). In the next step, $k$-SCC assigns each object in $D$ to its nearest cluster based on the dissimilarity function shown in Eq. (15). It then updates $k$ cluster center based on Eq. (8). The algorithm iterates the assignment and update steps until no further reassignment of objects takes place. The indexes of objects in $k$ clusters and the proximity matrix are used to compute the average silhouette value based on the Eq. (19) (line 10). The obtained average silhouette value is then put into the $SilSet$ (line 11). The $k$-SCC works in the same manner for every index $k$ until $k$ is larger than $k_{\max}$ (lines 3 to 11). Finally, the index $k$ that has the largest value of average silhouette in $SilSet$ is selected as the optimal index $k$, namely $k_{opt}$, for the given dataset. The algorithm then returns $k_{opt}$ clusters as the output of the algorithm (line 12).

## 5  Comparative Experiment

In this section, the proposed algorithm was tested on both real and synthetic datasets. The Car evaluation, Chess, Connect-4, Nursery, Soybean (small), Spect heart and Tictactoe are real datasets from the UCI repository[1]. The SD5K and SD10K are synthetic datasets generated by the Dataset Generator[2]. The characteristics of these datasets are shown in Table 3. The performance of $k$-SCC is compared with other three clustering algorithms: k-modes [8], Modified-3 [15] and an additional version of the $k$-SCC that uses the simple matching dissimilarity measure to compute the dissimilarity matrix in Algorithm 1, namely $k$-SCC+. All algorithms were implemented in Python using PyCharm. Experiments were performed on a high-performance VPCC cluster[3] equipped with an Intel Xeon Gold 6130 2.1 GHz (16 Cores×2), 64 GB of RAM, running CentOS 7.2 for each CPU node. The source code of $k$-SCC and datasets are provided at http://bit.ly/2HiwgKl. Since the four algorithms use random initialization to initially form centers of clusters, it may lead to non-repeatable clustering results. For that reason, each of the four algorithms was run 100 times for the initialization step and the best one was selected for each dataset.

**Table 3.** Characteristics of the experimental datasets

| Dataset | #instances | #attributes | #classes | Type |
|---|---|---|---|---|
| Car Evaluation | 1,728 | 6 | 4 | Real-life |
| Chess | 3,196 | 36 | 2 | Real-life |
| Connect-4 | 10,000 | 42 | 3 | Real-life |
| Nursery | 12,960 | 8 | 5 | Real-life |
| Soybean (small) | 47 | 35 | 4 | Real-life |
| Spect Heart | 267 | 22 | 2 | Real-life |
| Tic-Tac-Toe | 958 | 9 | 2 | Real-life |
| SD5K | 5,000 | 6 | 4 | Synthetic |
| SD10K | 10,000 | 6 | 2 | Synthetic |

---

[1] https://archive.ics.uci.edu/ml/index.php.
[2] http://www.datasetgenerator.com.
[3] https://www.jaist.ac.jp/iscenter/en/mpc/vpcc/.

### 5.1 Experimental Results

In the experiment, the average silhouette values were measured for the various number of clusters on each dataset. More specifically, $k$ is varied between two and ten for each dataset, then choose the best $k$ by comparing the average silhouette results obtained for the different $k$ values. Results are shown in Fig. 2. In this figure, vertical axes denote the average silhouette value, and horizontal axes indicate the number of clusters used. In general, $k$-SCC has higher average silhouette values than those of $k$-modes, Modified-3 and $k$-SCC+ for all datasets. On Car Evaluation, it can be classified into two or four clusters, which correspond to the largest and second largest average silhouettes for this dataset. On Chess, this dataset can be classified into two or three clusters based on its largest and second largest average silhouettes, respectively. Similar results can be observed for the other datasets. The results show that the average silhouette values of $k$-SCC and $k$-SCC+ are higher than those of $k$-modes. It means that the kernel density estimation used to form cluster centers used in $k$-SCC and $k$-SCC+ are more efficient than the modes used in $k$-modes. In addition, the information-theoretic based dissimilarity measure used in $k$-SCC and $k$-SCC+ are more efficient than the simple matching dissimilarity used in $k$-modes. It can be also observed that the average silhouettes of $k$-SCC are better than those of $k$-SCC+. Thus, the information-theoretic based dissimilarity measure used to compute the dissimilarity matrix is more effective than the simple matching used in $k$-SCC+. Furthermore, the average silhouette of $k$-SCC are better than those of the Modified-3 in most cases. It is worth to note that the runtime of Modified-3 is much higher than that of $k$-SCC because the former uses a feature weighting scheme to automatically measure the contribution of individual attributes for the cluster. Thus, $k$-SCC is more efficient in both estimation and computation than Modified-3. In general, the proposed framework is sensitive to the choice of dissimilarity measure to determine distances between data objects and the way to perform the clustering step.

### 5.2 A Case Study: Sake Wine Dataset

Another experiment was performed to evaluate the performance of compared algorithms on the Sake wine dataset, which is a real dataset collected from Fujinami lab[4]. More specifically, the taste sensing system named TS-5000Z, which is employed the same mechanism as that of the human tongue is used to convert the taste of Sake wine into numerical data. The TS-5000Z evaluates two types of taste: initial taste, which is the taste perceived when food first enters the mouth, and aftertaste, which is the persistent taste that remains in the mouth after the food has been swallowed. The initial taste is indicated by six features: *sourness*, *bitterness*, *astringency*, *umami*, *saltiness* and *sweetness*, while the aftertaste is indicated by *aftertaste from astringency* and *richness*. The dataset contains 68 Sake samples with eight features corresponding to the initial taste and aftertaste.

---

[4] http://www.jaist.ac.jp/~fuji/index.html.

Figure 3 shows a classification of the Sake data by using the complete-linkage hierarchical clustering. If the dendrogram is cut at the height of eight, nine or ten, then four (with one outlier), three or two clusters are obtained, respectively. The original Sake wine dataset obtained by using the TS-5000Z is a numeric data. We first pre-process the original data so as to be applicable to $k$-SCC. In this work, Kansei words with five linguistic grades: *very low*, *low*, *neither*, *high*, *very high* are used to convert the numeric values into categories in each feature. We first discretized each attribute to five scales corresponding to five linguistic grades. We then mapped each value in each attribute into the corresponding grade. The average silhouette values of $k$-SCC and compared algorithms are shown in Table 4. It is observed that the $k$-SCC outperforms $k$-modes, Modified-3 and $k$-SCC+ in most cases, except for $k = 10$ where the average silhouette of $k$-SCC+ is higher than one of $k$-SCC. Figure 4 shows silhouette plots of the Sake wine data for $k \in [2, 10]$. According to the results in Table 4, the Sake samples can be classified into three or four clusters, which correspond to the largest and second largest average silhouettes for this dataset. In other words, the recommended number of clusters match the results obtained in Fig. 3.

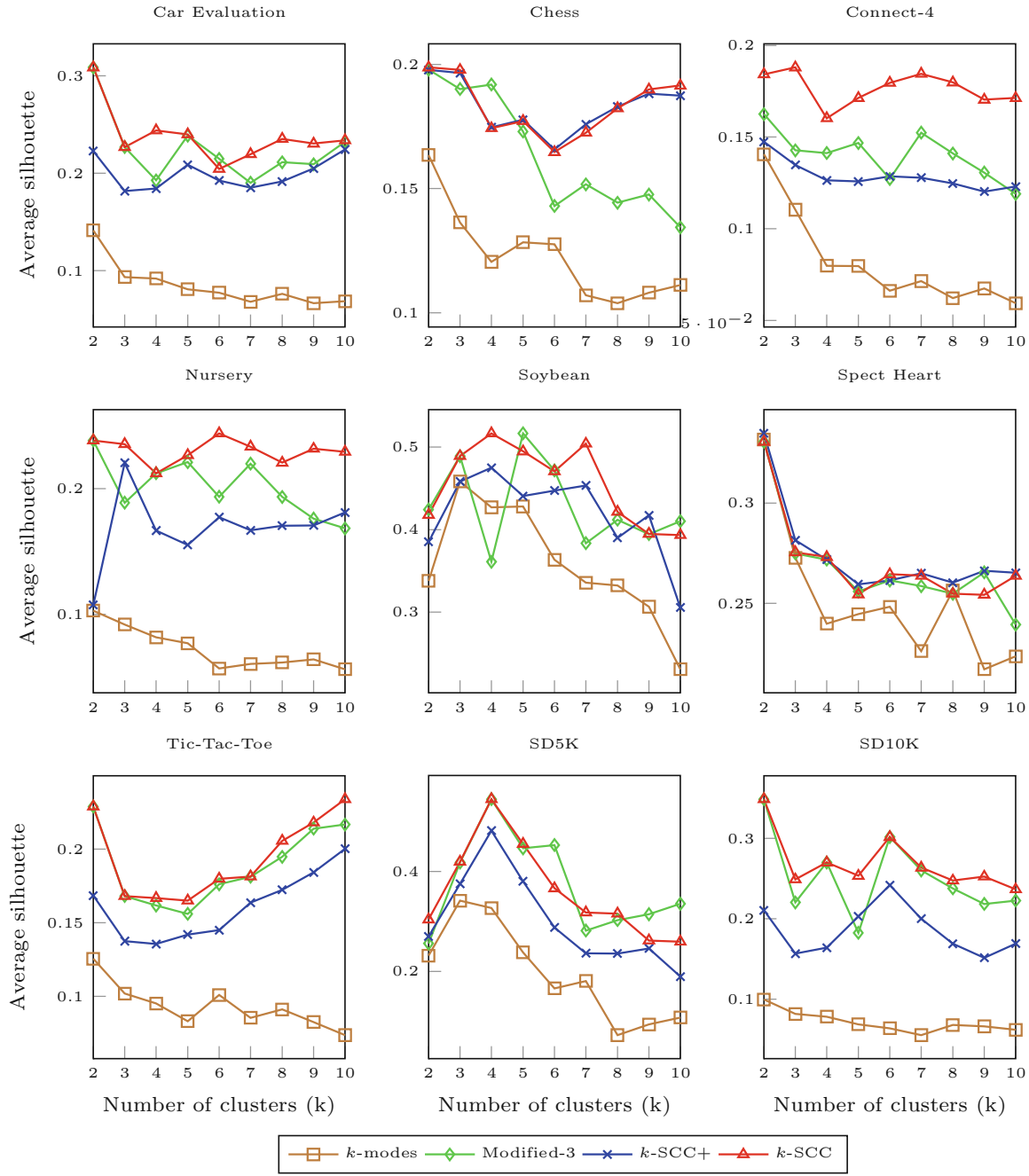

**Fig. 2.** Average silhouette for various number of clusters

## 6 Summary and Discussion

This paper has proposed an algorithm named $k$-SCC for estimating the optimal numbers of clusters in categorical data clustering. $K$-SCC uses the silhouette approach to find the optimal $k$ corresponding to the highest average silhouette value in each dataset. To perform the clustering steps, the proposed algorithm uses the kernel density estimation method and an information-theoretic based dissimilarity to form cluster centers and measure the distance between objects and cluster centers, respectively. This dissimilarity is also used to build the dissimilarity matrix when computing the average silhouette values. The experimental results have shown that the proposed algorithm outperforms the other three compared algorithms on both synthesis and real datasets.

The advantages of the proposed algorithm are as follows. First, this method can deal with categorical data, which is very popular in many real-life applications. In addition, the proposed method can be extended to estimate the number of clusters in mixed numeric and categorical data clustering. Second, the proposed method is a partitional clustering, which means that it will terminate at a local optimum. Third, the proposed method can be upgraded by using other dissimilarity measures to compute the dissimilarity matrix and distances between data objects and cluster centers.

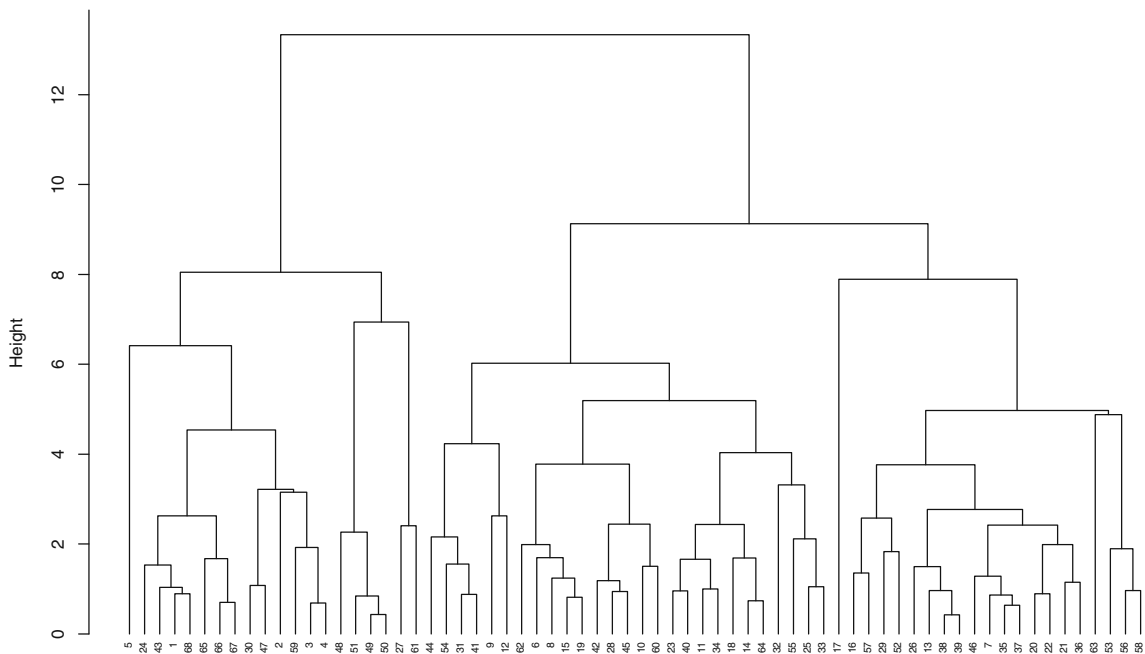

**Fig. 3.** Hierarchical clustering on Sake wine data using Complete-lingkage

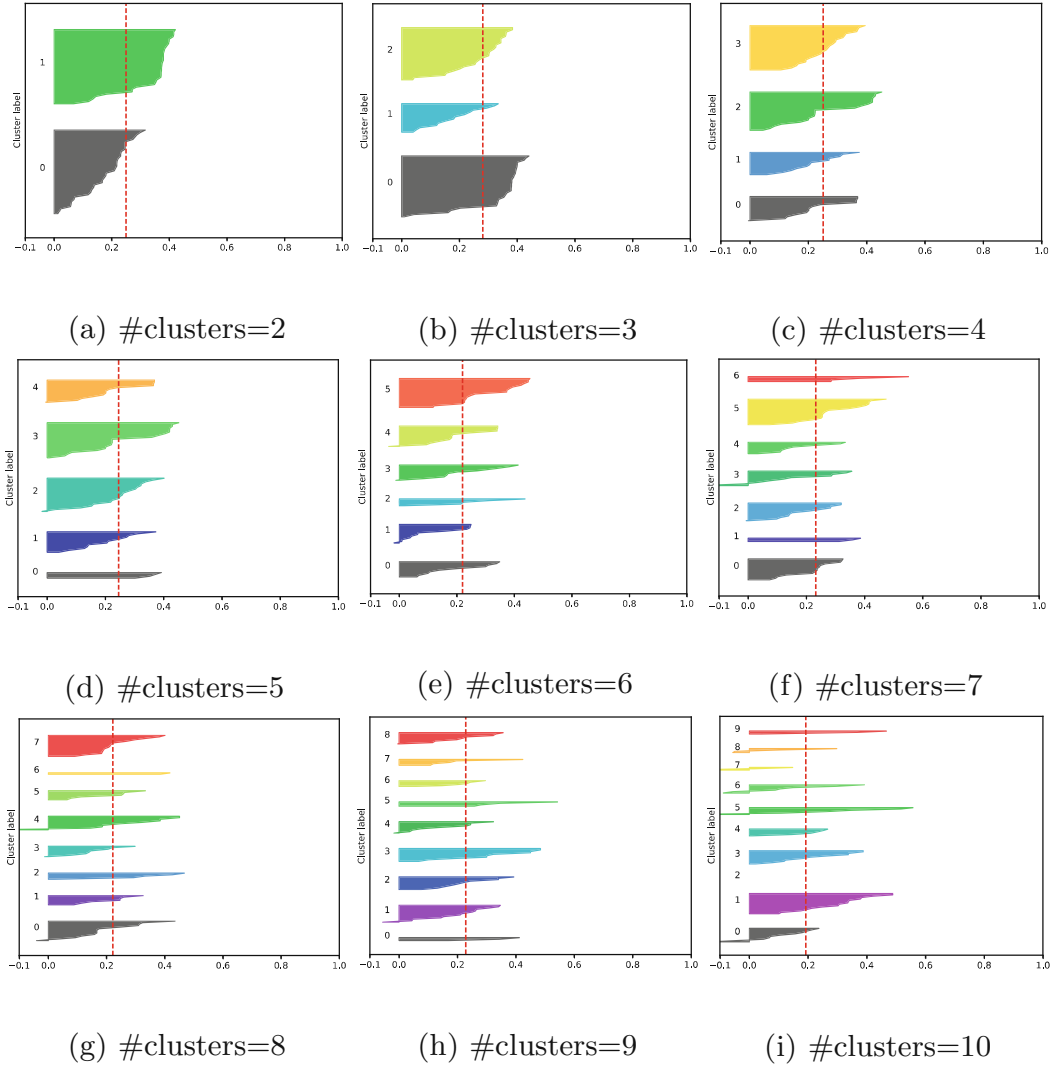

**Fig. 4.** Silhouette plots of Sake wine dataset for $k$ in range of $[2, 10]$

**Table 4.** Average silhouette values on the Sake wine dataset

| k | Algos | | | |
|---|---|---|---|---|
| | $k$-modes | Modified-3 | $k$-SCC+ | $k$-SCC |
| 2 | 0.1461 | **0.2501** | 0.2014 | **0.2501** |
| 3 | 0.1405 | 0.2744 | 0.2096 | **0.2811** |
| 4 | 0.1562 | 0.2252 | 0.1584 | **0.2585** |
| 5 | 0.1351 | **0.2511** | 0.1923 | 0.2451 |
| 6 | 0.1736 | 0.2135 | 0.1879 | **0.2200** |
| 7 | 0.1778 | **0.2556** | 0.2128 | 0.2328 |
| 8 | 0.1512 | 0.2028 | 0.2204 | **0.2217** |
| 9 | 0.1168 | 0.2168 | 0.1968 | **0.2288** |
| 10 | 0.1412 | 0.1469 | **0.2148** | 0.1924 |

The inherent limitations of the proposed algorithm are as follows. The average silhouette values strongly depend on the dissimilarity measure used to compute the pairwise dissimilarity matrix and the clustering scheme to partition the input datasets. In addition, the proposed method has high costs when applying for large-scale datasets. Thus, using a suitable clustering framework to get high silhouette values and reducing the computational complexity is necessary for this task.

In future work, we will extend the proposed approach to the problem of estimating the number of clusters in mixed data and design a parallel method to speed up the computational process.

**Acknowledgment.** This paper is based upon work supported in part by the Air Force Office of Scientific Research/Asian Office of Aerospace Research and Development (AFOSR/AOARD) under award number FA2386-17-1-4046.